\documentclass[lettersize,journal]{IEEEtran}
\usepackage{amsmath,amsfonts}
\usepackage{algorithmic}
\usepackage{algorithm}
\usepackage{array}
\usepackage[caption=false,font=normalsize,labelfont=sf,textfont=sf]{subfig}
\usepackage{textcomp}
\usepackage{stfloats}
\usepackage{url}
\usepackage{verbatim}
\usepackage{graphicx}
\usepackage{cite}
\usepackage{booktabs}
\usepackage{multirow}
\usepackage{times,latexsym}
\usepackage{url}
\usepackage[T1]{fontenc}
\usepackage{times}
\usepackage{latexsym}
\usepackage{booktabs}
\usepackage{multirow}
\usepackage{amsmath}
\usepackage{graphicx}
\usepackage{mdframed}
\usepackage{CJK}
\usepackage{subcaption}
\usepackage[utf8]{inputenc} 
\usepackage[T1]{fontenc}    
\usepackage{hyperref}       
\usepackage{url}            
\usepackage{booktabs}       
\usepackage{amsfonts}       
\usepackage{nicefrac}       
\usepackage{microtype}      
\usepackage{xcolor}         
\usepackage[misc,geometry]{ifsym}
\usepackage{tabularx}

\usepackage{subcaption}  
\usepackage{graphicx}    
\newcommand{\ph}{\phantom{0}}

\newcommand*{\escape}[1]{\texttt{\textbackslash#1}}

\usepackage[numbers]{natbib}

\begin{document}

\title{Benchmarking GPT-4 against Human Translators: A Comprehensive Evaluation Across Languages, Domains, and Expertise Levels}

\author{Jianhao Yan, Pingchuan Yan, Yulong Chen, Jing Li, Xianchao Zhu, and Yue Zhang
\thanks{
Jianhao Yan is with
Zhejiang University, Zhejiang, P.R. China, and also with the School of Engineering, Westlake University, Zhejiang, P.R. China (e-mail: elliottyan37@gmail.com).
}
\thanks{
Pingchuan Yan is with the University College London, London, UK (e-mail: ypc1956280693@gmail.com).
}
\thanks{
Yulong Chen is with the Department of Computer Science and Technology, University of Cambridge, Cambridge, UK.
}
\thanks{
Jing Li and Xianchao Zhu are with the LanBridge Group, Sichuan, P.R. China
}
\thanks{
Yue Zhang is with the School of Engineering, Westlake University, Zhejiang, P.R. China, and also with Institute of Advanced Technology, Westlake Institute for Advanced Study (e-mail: yue.zhang@wias.org.cn).
}
\thanks{Corresponding author: Yue Zhang.}}



\maketitle

\begin{abstract}
This study presents a comprehensive evaluation of GPT-4's translation capabilities compared to human translators of varying expertise levels. Through systematic human evaluation using the MQM schema, we assess translations across three language pairs (Chinese$\longleftrightarrow$English, Russian$\longleftrightarrow$English, and Chinese$\longleftrightarrow$Hindi) and three domains (News, Technology, and Biomedical). Our findings reveal that GPT-4 achieves performance comparable to junior-level translators in terms of total errors, while still lagging behind senior translators. Unlike traditional Neural Machine Translation systems, which show significant performance degradation in resource-poor language directions, GPT-4 maintains consistent translation quality across all evaluated language pairs. Through qualitative analysis, we identify distinctive patterns in translation approaches: GPT-4 tends toward overly literal translations and exhibits lexical inconsistency, while human translators sometimes over-interpret context and introduce hallucinations. This study represents the first systematic comparison between LLM and human translators across different proficiency levels, providing valuable insights into the current capabilities and limitations of LLM-based translation systems.

\end{abstract}


\section{Introduction}

\IEEEPARstart{R}{ecent} studies show that LLMs can serve as a strong translation system and a good substitute for NMT models~\cite{jiao2023chatgptgoodtranslatoryes, wang2023documentlevel, enis2024llm, huang2023towards, wu2024adapting, hendy2023good, peng2023towards}. For example, \cite{jiao2023chatgptgoodtranslatoryes} and \cite{wang2023documentlevel} find that GPT-4 can outperform commercial machine translation systems using both automatic and human evaluation. Such impressive results have fostered a wide range of applications, such as the use of GPT-4 for literary translation~\cite{wu2024perhaps}, poetry translation~\cite{poetry} and cross-lingual summarization~\cite{cross_summ}.

With the advance of LLM translation, traditional evaluation metrics such as BLEU are inadequate for evaluating their quality~\cite{chaganty-etal-2018-price,wmt19}.  While the research community has taken model-based approaches in recent work~\cite{comet,bleurt},  these metrics can suffer from model biases and distribution shifts. We take a different perspective by benchmarking them against human translators using systematic human evaluation.
Our goal is to understand LLM translators by putting them into the translation industry, hiring professionals to evaluate their translation quality against the translation quality of different levels of human translators, and gaining insight into any systematic differences between LLM translations and human translations~\cite{jiao2023chatgptgoodtranslatoryes}.
Such evaluation can complement existing studies and give a more comprehensive understanding of LLM translation quality.

To determine where LLMs fall within the spectrum of human translation proficiency, we take the current representative LLM, i.e., GPT-4, comparing it against human translators with different expertise, ranging from novice translators to seasoned professionals.
A preliminary study comparing human translations against GPT-4 translations shows that \emph{even experts cannot reach a high consensus on which translation is better}, despite that they tend to give higher scores to human translations.  
The results indicate that LLMs do generate translations that approach the human level,  and there can be potential subtle differences between LLMs and human translators, which are worth a more in-depth study.

Given these findings, we take a finer-grained evaluation across different languages and domains, so that translation quality can be better calibrated and systematic differences can be measured.
Our evaluation covers three language pairs from resource-rich to resource-poor, i.e., Chinese$\leftrightarrow$English, Russian$\leftrightarrow$English, and Chinese$\leftrightarrow$Hindi, and three domains, i.e., News, Technology, and Biomedicine. 
Given a source sentence, we ask junior, medium, senior human translators and two machine translators, i.e., GPT-4 and Seamless, a traditional NMT translator, to generate the corresponding translation in the target language. 
Then we hire independent expert annotators to label the errors in the target sentence under the MQM schema~\cite{freitag2021experts}.
Results show that while traditional NMT systems significantly underperform, GPT-4 reaches a comparable performance to junior/medium-level translators in the perspective of total errors made, yet lags behind senior ones with a considerable gap. GPT-4 represents a significant milestone in neural machine translation, establishing a baseline for legitimate comparative analysis with human translation performance.

Further language-specific results demonstrate that GPT-4 mitigates traditional machine translators' drawback of significant performance gap from resource-high to resource-low directions. 
Our traditional MT translator Seamless can be a reliable translator for English$\leftrightarrow$Russian, but shows poor quality for other language pairs. In contrast, GPT-4 demonstrates consistent performance across all evaluated language directions, achieving quality levels on par with junior to mid-level human translators.
Across various domains, similar comparisons of the GPT-4 translator against human translators are observed. 
Interestingly, each translator has its own drawbacks, while the GPT-4 translator is weak in Grammar and Named Entity, it does not suffer from human hallucination and fatigue.
More specifically, we observe that GPT-4 exhibits two primary limitations: adherence to overly literal translations and lexical inconsistency. Conversely, human translators tend to over-interpret certain contextual cues and introduce hallucinations.

To our knowledge, we are the first to evaluate LLMs against various levels of professional human translators and analyze the systematic differences between LLMs and human translators. \footnote{We release our code and data at \url{https://github.com/ElliottYan/GPT_versus_mt_experts}.}







\section{Related Work}
\paragraph{Benchmarking LLMs}
Previous studies have benchmarked LLMs on various NLP tasks. 
\cite{xu2020clue} benchmark several LLMs on Chinese text, evaluating their Chinese ability.
\cite{ye2024benchmarking} assess LLMs through Question Answering (QA), MMLU \cite{hendrycks2021measuring}, and other metrics. From these tests, LLMs with larger scales are generally proved to be more accurate except for certain tasks. \cite{yuan2023revisiting} demonstrates that LLMs perform well in long-context understanding and are more capable with Out-of-Distribution, which means LLMs have a certain degree of generalization ability. 

Further to the MT field, \cite{jiao2023chatgpt} find that GPT-4 performed competitively with other SotA translation products. \cite{wang2023documentlevel} further investigated the capability of GPT-4 in document-level translation, the results show that GPT-4 performs better than commercial translation products and document NMT methods.
Compared to them, our work empirically shows that GPT-4 is comparable to junior human translators.

\paragraph{LLMs as Human Experts}
Due to the great capacities of GPT-4 over traditional NLP models, researchers have investigated and compared the performance of GPT-4 as human experts in multiple NLP tasks.
\cite{zhu2024benchmarking} highlight that GPT-4 and GPT-4-turbo show top performance on a Chinese financial language understanding task. 
\cite{liu2023evaluating} find the LLMs can be beneficial to biomedical NLP tasks.
\cite{goyal2022news} compare GPT models with several summarization models and humans, and find that GPT can generate summaries preferred by humans.
In AI for education area, \cite{nguyen2024using} show GPT-4's can provide teaching feedback for students. 
\cite{maloney2024comparison} find that GPT-4 shows close performance compared with human participants in coordination games.
\cite{siu2023chatgpt} show that GPT-4 is comparable to humans on technical translation tasks.
\cite{bojic2023gpt} find that GPT-4 can outperform human experts on linguistic pragmatic tasks. 
In clinical diagnostics, \cite{han2023comparative} find that GPT-4 can give comparable performance to humans, and GPT-4v (vision version) can even outperform human experts.

\paragraph{Human Evaluation for MT}
\cite{da} first propose Direct Assessment~(DA), which uses a continuous score from 0 to 100 to represent the quality of a hypothesis. 
DA has been adopted in WMT translation tasks for the past few years~\cite{wmt21,wmt22,wmt23}. 
MQM~\cite{lommel2014assessing}, the annotation used in this paper, is another widely used annotation scheme~\cite{mqm-usage-1,mqm-usage-2}. 
It requires the annotators to annotate the error span for each hypothesis and is shown to be more accurate and reliable than DA~\cite{freitag2021experts}. 
Thus, it is utilized in the metrics tasks of 2022 and 2023 WMT challenges~\cite{wmt22-metrics,wmt23-metrics}. 

\paragraph{Human Parity}
The human parity for machine translation systems is first claimed by \cite{hassan2018achieving}, which describes a comparable performance on the WMT 2017 news translation task from Chinese to English when compared to professional human translations. 
However, this claim is challenged by the following research, raising concerns about the limited scope of human parity. 
These limitations include the expertise of human evaluators~\citep{fischer-laubli-2020-whats}, the origin and quality of source sentences~\citep{toral-etal-2018-attaining,202110.0199}, the limited scenario of comparison~\citep{poibeau2022human} and difficulty of translation~\citep{graham-etal-2020-assessing}, indicating significant gaps between NMT models and the professional translators. 
In this work, we evaluate whether the SOTA LLM GPT-4 performs comparable to professional translators and what differs between human translators and LLMs. 
With the above lessons in mind, we address these limitations by hiring expert annotators, avoiding target-origin source text, manually evaluating source sentences, and covering high-resource to low-resource language pairs and various domains. 




\section{Preliminary Study}\label{sec:pre}
We aim to first compare GPT-4 translations with human translations qualitatively, in a coarse manner. 
We hire expert annotators from the Lan-Bridge Group\footnote{\url{https://www.lan-bridge.co.uk/}}, who are native Chinese speakers and dedicated to Chinese-English translation for over 5 years. 
Our comparison is simple and direct. We sample human-translated texts and prompt GPT-4 to translate the same source sentence. Then, we ask expert annotators to determine which translation is better. 

Particularly, to have a quick overview of the qualities of human translations against GPT-4 translations, we first utilize COMET-QE\footnote{Unbabel/wmt23-cometkiwi-da-xl} to score our in-house Chinese to English human-translated documents, and select two documents with the highest score and the lowest score. Note that our in-house translated documents are all translated by professional translators. 
In this way, we gather 40 pairs of translations from professional translators and GPT-4, respectively. 
Recent findings~\cite{freitag2021experts} have demonstrated that crowd-sourced human ratings are less reliable for high-quality MT evaluation. 
Thus, we hire six expert annotators to compare the two translations and select the better translations they find. 
We randomly shuffle the GPT-4 and human translations to prevent annotators from identifying GPT-4. 

The average win rate of GPT is 15.5/40~(36.25\%), which indicates a clear win for human translators. However, when delving deeper, we find that the expert annotators have a low ratio of agreement with each other. 
In Table \ref{tab:compare_kappa}, the inter-agreement ratio between most annotators is around 60\%~(the baseline is 50\%). 
A significance test shows that only annotator B finds human translation significantly better than GPT's translation, while the other annotators have high p-values. 
These results indicate that \textit{even expert annotators find it difficult to agree on which translation is better}, and GPT-generated translations might have different advantages against human-generated ones. 
We are therefore motivated to conduct a finer-grained and comprehensive evaluation to reveal the systematic difference between GPT-4 and human translations. 


\begin{table}[]
\centering
{
\begin{tabular}{c|cccccc}
\toprule
Annotators & A     & B     & C     & D     & E     & F     
\\\midrule
A          & 100.0 & 57.5  & 65.0  & 65.0  & 62.5  & 67.5  \\
B          &    -   & 100.0 & 52.5  & 52.5  & 50.0  & 50.0  \\
C          &   -    &    -   & 100.0 & 65.0  & 82.5  & 67.5  \\
D          &   -    &    -   &   -    & 100.0 & 57.5  & 62.5  \\
E          &     -  &   -    &    -   &    -   & 100.0 & 70.0  \\
F          &  -     &    -   &      - &     -  &   -    & 100.0 \\\midrule
p-value & 1.000 & 0.038 & 0.268 & 0.081 & 0.154 & 0.875 \\\bottomrule
\end{tabular}}
\caption{Ratio(\%) of agreed winner across expert annotators and significance p-value for binomial test. P-value < 0.05 denotes a significant difference between GPT-4 and Human.}
\label{tab:compare_kappa}\end{table}

\renewcommand{\arraystretch}{1.2}
\newcommand{\centered}[1]{\begin{tabular}{l} #1 \end{tabular}}

\begin{table*}[]
\centering
\small
\begin{tabular}{|c|l|p{5cm}|p{5cm}|}
\hline
\textbf{Type}                     & \textbf{Error Name}                 & \texttt{\textbf{Explanations}} & \textit{\textbf{Examples}}                  \\ \hline
\multirow{6}{*}{Accuracy} & Mistranslation   & \texttt{Translation does not accurately represent the source.} & {                        \emph{[EN] It has to be done by the book.} 
                        
                        \emph{[FR] Il doit \^{e}tre fait [par le livre]MISTRANSLATION} 
}             \\ \cline{2-4} 
                         & Addition                   & \texttt{Information not present in the source.} &{\emph{[EN] That way you can be sure that you were the one who made the changes.} 
                        
                        \emph{[ES] As\'{i} puedes estar seguro de que fuiste t\'{u} quien hizo [todos]ADDITION los cambios.} 
                      }               \\ \cline{2-4} 
                          & MT Hallucination & \texttt{Information that has nothing related to source; or gibberish; or repeats } & {\emph{[EN] You can send us a follow-up email at this address [EMAIL].} 
                        
                        \emph{[ES] H\'{a}game saber si tiene alguna otra pregunta]MT HALLUCINATION.}}                        \\ \cline{2-4} 
                         & Omission                   & \texttt{Missing content from the source. }  &   {\emph{[EN] We do not have much information on this.} 
                        
                        \emph{[FR] Nous ne disposons pas []OMISSION beaucoup d'informations \`{a} ce sujet.} }\\ \cline{2-4} 
                         & Untranslated               & \texttt{Not translated.}          & {\emph{[EN] How To Make Pizza Dough} 
                        
                        \emph{[FR] Comment faire de [Pizza Dough]UNTRANSLATED} }       \\ \cline{2-4} 
                         & Incorrect NE and Term & \texttt{Incorrect usage of NE and Terminology.} &{                        \emph{[EN] Dear Wiley, } 
                        
                        \emph{[IT] Gentile [Wilar]Incorrect NE, } }                   \\ \hline
\multirow{6}{*}{Fluency} & Grammar                    & \texttt{Problems with grammar of target language.} & { \emph{[EN] I understand that you want to check in online.} 
                    
                    \emph{[CS] ch\`{a}pu, ze se chcete [odbaven\'{i}]GRAMMAR online.} }  \\ \cline{2-4} 
                         & Punctuation                & \texttt{Incorrect punctuation (for locale or style)}    &  {\emph{[EN] Original copy of the Proof of Purchase or Invoice (not a screenshot): } 
                    
                    \emph{[PT] C'{o}pia original do comprovante de compra ou nota fiscal (n\~{a}o uma captura de tela)[.]PUNCTUATION} }   \\ \cline{2-4} 
                         & Spelling & \texttt{Incorrect spelling or capitalization.}&{\emph{[EN] [...], but we will seek assistance from higher support and see what we can do regarding this issue.} 
                    
                    \emph{[IT] [...], ma chieder\`{o} comunque aiuto ai responsabili dell'assistenza per capire che cosa [Zi]SPELLING pu\`{o} fare per quanto riguarda questo problema.} }\\ \cline{2-4} 
                          & Register & \texttt{Incorrect grammatical register (e.g., inappropriately informal pronouns).}  & {\emph{[EN] Wishing you a great day ahead.} 
                    \emph{[DE] Ich w\"{u}nsche [Ihnen]REGISTER einen sch\"{o}nen Tag.} }  \\ \cline{2-4} 
                         & Inconsistent Style & \texttt{Internal inconsistency ( not related to terminology )} & {\emph{[EN] Please click on this link. [...] This link will expire in 24 hours.} 
                
                    \emph{[NN] Klikk p\r{a} denne [lenken].[...]Denne [linken]INCONSISTENCY utloper om 24 timer.}}\\ \cline{2-4} 
                         & Unnatural Flow             & \texttt{Translations that are too literal or sound unnatural.} &{                    \emph{[EN] Zebras are ideal for animal matching.} 
                    
                    \emph{[DE] [Zebras sind ideal, um bestimmte Tiere zu finden]UNNATURAL FLOW.} 
 } \\ \hline
Other                    & Non-translation            & -  & -                                                   \\ \hline
\end{tabular}
\caption{Error category and explanations. We mainly follow the guidelines from Unbabel, and merge some errors to reduce the efforts for annotators to understand the annotation system.}
\label{tab:error_categories}
\end{table*}

\section{Main Experimental Setup}
We conduct a comprehensive and fine-grained evaluation of GPT-4 against professional human translators. 
Specifically, we employed the widely recognized Multidimensional Quality Metrics (MQM) framework~\cite{lommel2014assessing} and compared human translators with varying levels of expertise to GPT-4.
Our evaluation spans multiple languages and domains, aiming to furnish broad insights into these comparisons.


\subsection{Data Collection}

We strategically collect multilingual and multi-domain source sentences to evaluate translation across diverse resource levels. Based on GPT-3 corpus reports, we select four languages: English (dominant in most models), Russian~(0.1884\% words, 8th), Chinese~(0.099\% words, 16th), and Hindi~(0.00483\% words, 41st). This choice allows us to assess translations from resource-high to resource-low scenarios. 
English and Russian, both Indo-European, represent resource-high directions, while English and Chinese, from different language families, serve as resource-medium directions. 
Recognizing the scarcity of expert translators in low-resource languages, we use Chinese $\leftrightarrow$ Hindi as our resource-low directions. 
In summary, we evaluate six crucial directions: English $\leftrightarrow$ Chinese, English $\leftrightarrow$ Russian, and English $\leftrightarrow$ Hindi. This selection enables us to assess bidirectional translation performance across varied resource levels, providing valuable insights into current capability of the GPT4 translator against human translators.

For general domain Chinese$\Leftrightarrow$English and English$\Leftrightarrow$Russian translation, we sample source sentences from the test sets of WMT2023 and WMT2022, respectively. 
For Chinese$\Leftrightarrow$Hindi, we extract source news text from public websites.
For multi-domain evaluation data, we choose two domains, i.e., biomedicine and technology, evaluating Chinese to English translation. 
The source sentences are extracted news texts from public websites. 
We ensure that all sources are source language origin to avoid the effect of translationese.
We manually evaluate all source sentences for these tasks to ensure that the source sentences are not too easy or too short. 
Finally, each task contains 200 sentences, making our evaluation a total of 1600 sentences. 


\subsection{Human and Machine Translators}
We ask different human translators from the same company as Section \ref{sec:pre} to translate our source sentences into the target language. 
Translators are of three different levels of expertise, categorized into junior-level, medium-level, and senior-level translators. 
The level of expertise is ranked by in-house criteria covering the translators' educational background, translation experience, and practical proficiency. 
For instance, a junior translator could be a graduate student majoring in target languages and having one or two years of experience in the translation industry. A medium translator could be a translator who has 3-5 years of experience or is a native speaker of the target language. 
To be classified as a senior-level translator, an individual must possess a minimum of ten years of translation experience, demonstrate exceptional proficiency by achieving a score of 99\% on quarterly assessments, and hold the distinguished CATTI++ translation certification. 
This categorization follows the tradition of the human translation industry. 

For all directions except Zh-Hi and Hi-Zh, we collect three human translation results from each level of expertise. For Zh-Hi and Hi-Zh, we only have medium-level and senior-level translators due to the scarcity of translators. 

In practice, we find that current human translators also heavily rely on GPTs or machine translators as assistance. 
Thus, to enable a fair comparison, we prohibit human translators from using machine translation or GPTs as assistance.


We use \texttt{gpt-4-1106-preview}, the current state-of-the-art large language model released by OpenAI and \texttt{Seamless M4T}~\cite{seamless2023} as the representative of traditional machine translations to complement our experiments.
We directly prompt GPT-4 to obtain the translation, as it is the most common practice for normal users, the easiest to reproduce, and to avoid confusion by various techniques.

For GPT-4, we use greedy search for decoding, to ensure the reproducibility of the results. For SeamlessM4T, we use the 2.3B version of \texttt{seamlessM4T\_v2\_large} and adopt beam search with beam size 5. 

\begin{table*}[t]
\centering
\begin{tabular}{p{10cm}|p{1cm}}
\hline
\textbf{Prompt} & \textbf{COMET} \\\hline
Please translate the following sentence from Chinese into English. Your language and style should align with the language conventions of a native speaker. \escape{n}\textit{\{SOURCE\}}\escape{n}  & 0.775 \\\hline
You are an expert translator for translating Chinese to English. Your language and style should align with the language conventions of a native speaker. \escape{n}[Chinese]: \textit{\{SOURCE\}}\escape{n}[English]:& 0.755 \\\hline 
Please provide the English translation for these sentences. Your language and style should align with the language conventions of a native speaker. \escape{n}\textit{\{SOURCE\}}\escape{n} & \textbf{0.780} \\ 
\hline
\end{tabular}
\caption{Taking Chinese to English as an example, our three prompts and corresponding scores with COMET-QE. \textit{\{SOURCE\}} represents the source sentence to be translated. }
\label{tab:prompt_search}
\end{table*}

\subsection{Prompt Search}
Previous study~\cite{zhao2021calibrate,liu2023pre} shows that different prompts with LLMs can result in distinctive performance. 
Thus, we collect three candidate prompts used in previous research~\cite{xu2023paradigm,jiao2023chatgptgoodtranslatoryes} and use COMET-QE~\cite{rei2020comet} to select the best prompt to make the best use of GPT-4, as shown in Table \ref{tab:prompt_search}.
In particular, we use these three prompts to prompt GPT-4 to translate 100 source sentences in our Chinese-to-English test set and adopt COMET-QE to evaluate the quality of translations. 
We find that the third prompt yields the best performance, and hence we adopt this prompt for all following experiments. 

\subsection{Annotation Protocol}
To evaluate the results of candidates' systems, we hire experts to annotate the errors of translations blindly. 
The annotation platform is Doccano~\cite{doccano}, and the error tags are made according to MQM standards. 
MQM requires the annotators to annotate the span of errors in each hypothesis. 
All hypotheses of the same source sentence are shown to the annotator together to help decide which is better. 
We have 13 error categories and two severities, as shown in Table \ref{tab:error_categories}.
Our categorization for errors mostly follows Unbabel's practice~\footnote{\url{https://help.unbabel.com/hc/en-us/articles/6444304419479-Annotation-Guidelines-Typology-3-0}} and we focus on most common error types. 
Each tag has subtags with two severities, i.e., Minor or Major.
A screenshot of the annotation system is given in Figure \ref{fig:screenshot}.

For each task, we first ask the two expert annotators to carefully read our manual and conduct a training round on the first 10 groups of translations. 
Then, we manually check these annotations to provide feedback and ask the two annotators to check their disagreements and revise their results. 
After two rounds of such training processes, we ask the annotators to finish the remaining sentences without knowing each other's results. 

After the first round of annotation, we conduct a second round to further refine the evaluation results. In particular, we hire another two experts for each task and show them the previous annotation results. They are asked to approve and make necessary modifications to previous round annotations. 
\begin{table}[t]
\centering
{
\begin{tabular}{lrc}
\hline
\multicolumn{1}{|l|}{Task}  & \multicolumn{1}{l|}{Cohen Kappa(Segment)} & \multicolumn{1}{l|}{Krippendorffs(Span)} \\ \hline
\multicolumn{3}{|c|}{Reference, Re-Annotated by \cite{freitag2021experts}}\\ \hline
\multicolumn{1}{|l|}{WMT 2020 En-De}   & \multicolumn{1}{r|}{0.208}     &  \multicolumn{1}{r|}{0.456} \\ \hline
\multicolumn{1}{|l|}{WMT 2021 En-De}   & \multicolumn{1}{r|}{0.230}    &  \multicolumn{1}{r|}{0.501}    \\ \hline
\multicolumn{3}{|c|}{Ours}\\ \hline
\multicolumn{1}{|l|}{General Zh-En}    & \multicolumn{1}{r|}{0.257}    &  \multicolumn{1}{r|}{0.436}   \\ \hline
\multicolumn{1}{|l|}{General En-Zh}    & \multicolumn{1}{r|}{0.544}    &  \multicolumn{1}{r|}{0.579}   \\ \hline
\multicolumn{1}{|l|}{General En-Ru}    & \multicolumn{1}{r|}{0.461}    &  \multicolumn{1}{r|}{0.566}   \\ \hline
\multicolumn{1}{|l|}{General Ru-En}    & \multicolumn{1}{r|}{0.341}    &  \multicolumn{1}{r|}{0.875}   \\ \hline
\multicolumn{1}{|l|}{General Zh-Hi}    & \multicolumn{1}{r|}{0.256}    &  \multicolumn{1}{r|}{0.443}        \\ \hline
\multicolumn{1}{|l|}{General Hi-Zh}    & \multicolumn{1}{r|}{0.234}    &  \multicolumn{1}{r|}{0.495}        \\ \hline
\multicolumn{1}{|l|}{Technology Zh-En} & \multicolumn{1}{r|}{0.306}    &  \multicolumn{1}{r|}{0.581}   \\ \hline
\multicolumn{1}{|l|}{Biomedicine Zh-En} & \multicolumn{1}{r|}{0.373}    &  \multicolumn{1}{r|}{{0.616}}   \\ \hline
\multicolumn{1}{|l|}{\textbf{Average}} & \multicolumn{1}{r|}{\textbf{0.321}}    &  \multicolumn{1}{r|}{\textbf{0.555}}   \\ \hline
\end{tabular}}
\caption{Cohen Kappa (segment-level) and Krippendorffs' Alpha (span-level) agreement of our annotations. }
\label{tab:final_iaa}
\end{table}

\subsection{Inter-Annotator Agreement}


Error annotation with MQM is challenging, and previous work demonstrates that the agreement scores between MQM annotations are relatively low~\cite{lommel2014assessing}. Reasons for this could be disagreement on precise spans and ambiguous error categorization~\cite{lommel2014assessing}. Despite the low agreement scores, MQM is more reliable than other evaluation protocols like Direct Assessment~\cite{freitag2021experts}.

To compute inter-annotator agreement for MQM, we employ segment-level Cohen's Kappa~\cite{cohen1960coefficient} and span-level Krippendorff's alpha~\citep{krippendorff1980validity}. For reference, we calculate the agreement on the annotated results of the 2020 and 2021 WMT English-to-German tasks by~\cite{freitag2021experts}. 
Our IAA results are shown in Table \ref{tab:final_iaa}. 
Thanks to our two-round annotation process, our IAA scores show a favorable agreement, indicating a good annotation quality. 



\section{Main Results}
\subsection{Overall Results}
The overall results are shown in Table \ref{tab:error_number}, where each row represents a translator and each column represents the error score of each type. 
We compare the traditional machine translator, i.e., Seamless, LLM translator, i.e., GPT-4, and different levels of human translators. 

\noindent\paragraph{Error Severity}

The left part of Table \ref{tab:error_number} presents the averaged number of errors of different systems and translators. 
We normalize each error by the ratio of its span length against the length of the candidate translation.
Our results demonstrate that GPT-4 represents a significant milestone in neural machine translation, establishing a baseline for legitimate comparative analysis with human translation performance, while previous models demonstrably lag behind human capabilities.
Specifically, compared to our MT baseline~(seamless), GPT-4 has significantly fewer errors, i.e., 29.52 vs 20.43 minor errors and 17.35 vs 3.71 major errors.
\emph{GPT4 translator is comparable with junior/medium-level human translators}, producing 17.35/3.71 minor/major errors compared to 18.19/3.27 for junior-level and 20.19/3.30 for medium-level translators.
However, GPT4 lags behind senior translators by a significant margin, i.e., 3.71 vs 1.83 major errors, indicating machine translation is \textit{yet a solved problem}. 
To our knowledge, this is the first report on GPT-4 as a translator against various expertise levels of human translators. 


\begin{table}[t]
\centering
\begin{tabular}{@{}|c|c|c|c|c|c|@{}}
\hline
\multicolumn{2}{|l|}{}                             & \multicolumn{2}{c|}{\textbf{Severity}}          & \multicolumn{2}{c|}{\textbf{Error Type}} \\\cline{3-6}
\multicolumn{2}{|c|}{\multirow{-2}{*}{\textbf{Translator}}} & Minor                         & Major & Accuracy       & Fluency       \\\hline
                                & Seamless       & 29.52&17.35	& 28.25 & 17.37 \\\cline{2-6}
\multirow{-2}{*}{\textbf{Machine}}       & GPT4           & 20.43&\phantom{0}3.71& 11.12 & 12.95 \\\hline
        & Junior         & 18.19 & \phantom{0}3.27	 & \phantom{0}8.55 & 12.74        \\\cline{2-6}
        & Medium         & 20.19 & \phantom{0}3.30 &	12.58	& 10.66          \\\cline{2-6}
\multirow{-3}{*}{\textbf{Human}} & Senior & 12.04&\phantom{0}1.83	&\phantom{0}7.93 & \phantom{0}5.93 \\\hline
\end{tabular}
\caption{Overall Results: Average score of errors by different translators.}\label{tab:error_number}
\end{table}

\begin{figure*}[t]
    \centering
    \includegraphics[width=0.70\linewidth]{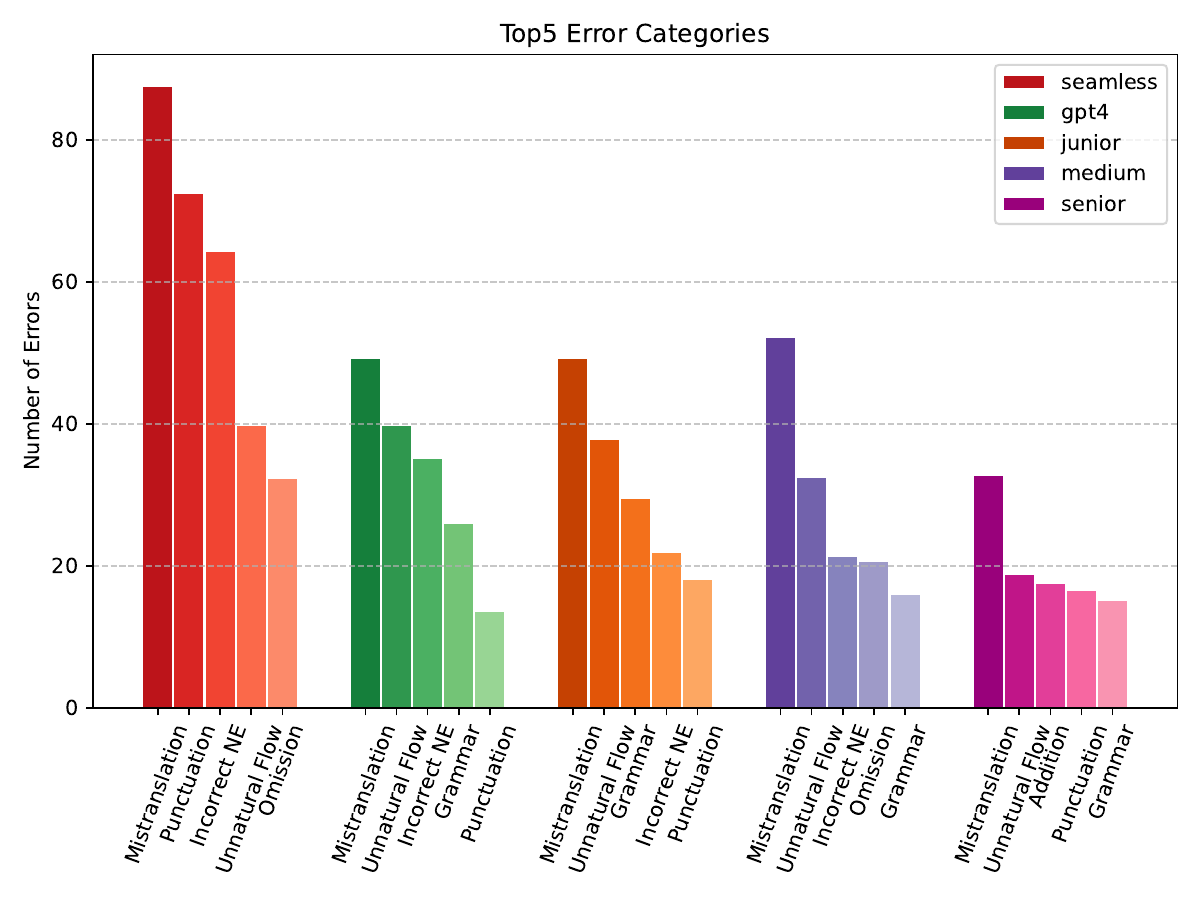}
    \caption{Top 5 categories of errors made by each translator.}
    \label{fig:top5_cat}
\end{figure*}

\paragraph{Error Categories}
\label{sec:overall}
The right part of Table \ref{tab:error_number} shows our overall results for two categories of MQM errors, namely Accuracy and Fluency~\cite{freitag2021experts}. We merge different error severity by assigning them weights, where a minor error weights 1 and a major error weights 3, as done in \cite{freitag2021experts}.
GPT-4's translation demonstrates good accuracy but worse language usage when compared with expert translators.
In particular, GPT-4 makes comparable errors in the accuracy of translations with junior/medium human translators~(11.12 vs 8.55/12.58), while it makes 12.95 fluency errors, weaker than all human translators.
We also find the seniority of human translators mainly manifests progressively refined language usage, i.e., fluency errors from 12.74 to 5.93, while accuracy errors show fluctuations across seniority levels.

We further investigate errors made by each translator. 
Figure \ref{fig:top5_cat} shows the top 5 categories of errors made by different systems. 
`Mistranslation' is the most frequent error made by all systems. Improving much over the seamless baseline, GPT-4 makes comparable numbers of `Mistranslation' with junior and medium human translators. 
`Unnatural Flow' is among the most frequent errors for all translators. 
Seamless, GPT-4, and junior translators have similar levels of `Unnatural Flow', indicating possible issues of literal translation and not following language conventions. 
In contrast, medium and senior translators are annotated with significantly fewer `Unnatural Flow' errors. 

GPT-4 makes much fewer `Incorrect Named Entity(NE)' errors compared to Seamless, which can be because of the huge knowledge acquired in the pre-training stage. However, it still has a gap compared to human translators. 
Finally, we notice that GPT-4 does not have Omission or Addition problems in its top-5 errors, whereas even senior translators have Addition errors. 
This can be because of the tendency for machine translators to use literal translation. 
We will discuss this further in the following sections. 


\subsection{Language-Specific Results}
\begin{figure*}[t]
    \centering
    \begin{minipage}{0.32\textwidth}
        \centering
        \includegraphics[width=\textwidth]{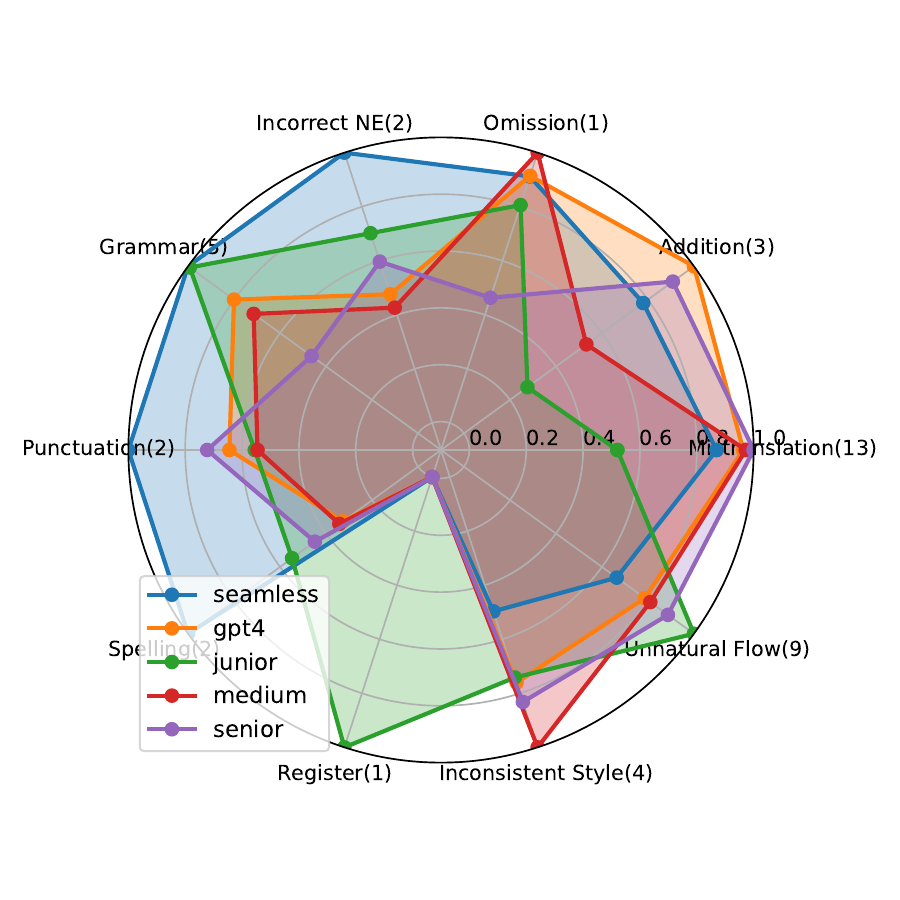}
        \caption*{(a) English$\leftrightarrow$Russian}
    \end{minipage}
    \hfill
    \begin{minipage}{0.32\textwidth}
        \centering
        \includegraphics[width=\textwidth]{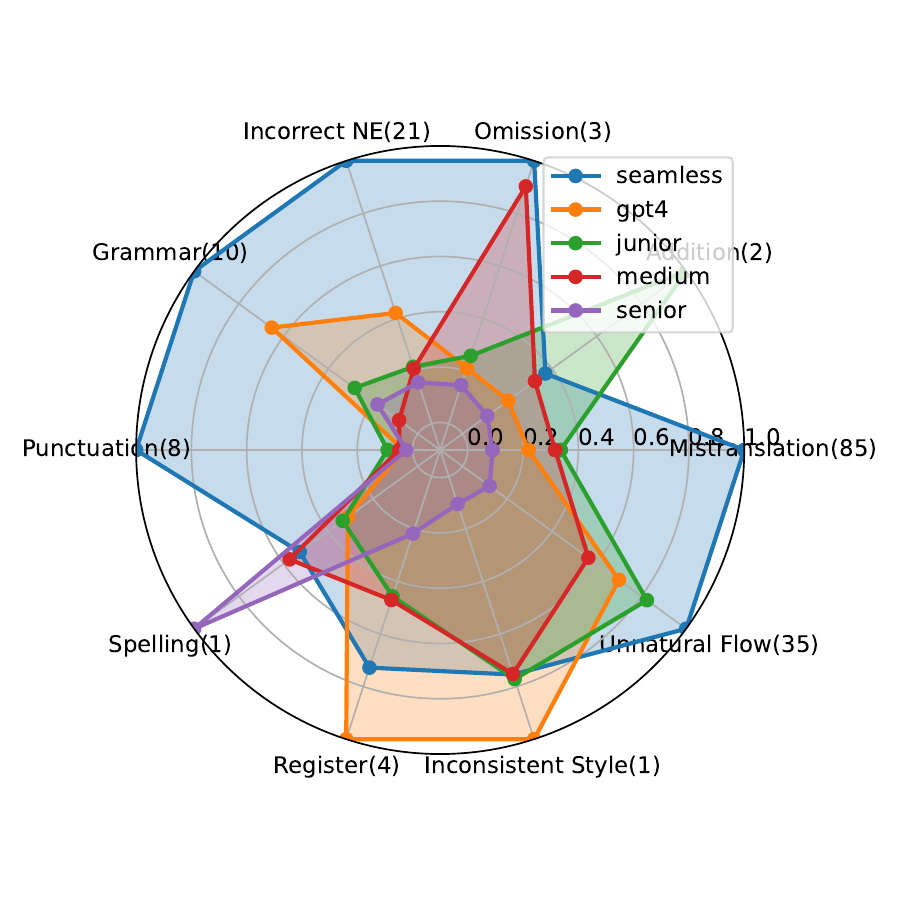}
        \caption*{(b) Chinese$\leftrightarrow$English}
    \end{minipage}
    \hfill
    \begin{minipage}{0.32\textwidth}
        \centering
        \includegraphics[width=\textwidth]{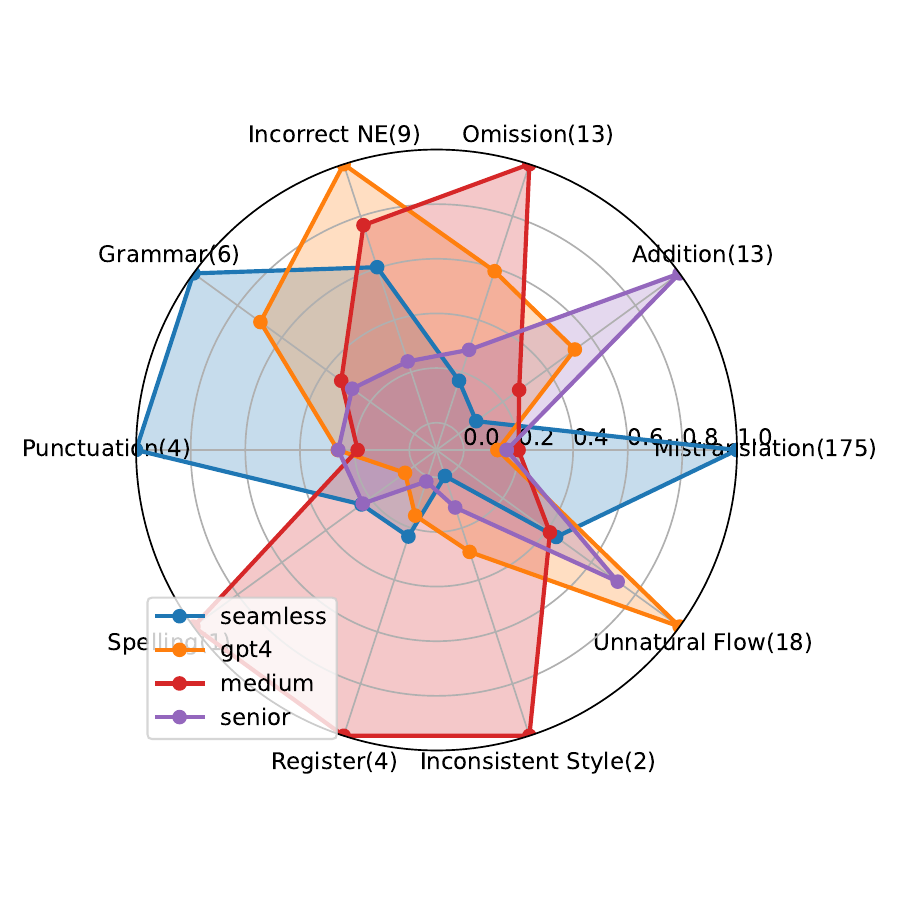}
        \caption*{(c) Chinese$\leftrightarrow$Hindi}
    \end{minipage}
    \caption{Detailed errors of categories for each domain. Higher values indicate more errors and the number after each error type is the maximum number of that error.}
    \label{fig:radar_lang}
\end{figure*}

We delve deeper into the detailed results for different languages, domains, and cases. The language-specific results are shown in Table \ref{tab:lang_spec}, which provides our language-specific results on Accuracy and Fluency, as in the previous section, average to and from directions of a language pair.
We give language-specific discussions below, with more detailed metrics being plotted in Figure \ref{fig:radar_lang}
.

\begin{table*}[t]
\centering
\begin{tabular}{|c|c|c|c|c|c|c|c|}
\hline
\multicolumn{2}{|c|}{\multirow{2}{*}{\textbf{Translator}}} & \multicolumn{2}{c|}{\textbf{En$\leftrightarrow$Ru}} & \multicolumn{2}{c|}{\textbf{Zh$\leftrightarrow$En}} & \multicolumn{2}{c|}{\textbf{Zh$\leftrightarrow$Hi}} \\\cline{3-8}
\multicolumn{2}{|c|}{}                & Accuracy & Fluency & Accuracy & Fluency & Accuracy & Fluency \\\hline
\multirow{2}{*}{\textbf{Machine}} & Seamless & 14.79    & 14.60   & 129.19   & 55.46   & 202.92   & 17.60   \\\cline{2-8}
                         & GPT4     & 15.82    & 14.11   & 29.57    & 34.98   & 44.36    & 22.83   \\\cline{1-8}
\multirow{3}{*}{\textbf{Human}}   & Junior   & 8.52     & 17.14   & 35.99    & 34.18   & -    & -   \\\cline{2-8}
                         & Medium   & 15.09    & 14.79   & 37.59    & 23.04   & 59.03    & 22.83   \\\cline{2-8}
                         & Senior   & 15.81    & 14.16   & 12.40    & 7.60    & 45.39    & 15.19  \\\hline
\end{tabular}
\caption{Language specific results.}\label{tab:lang_spec}
\end{table*}


\paragraph{English$\leftrightarrow$Russian} 
For English-Russian translation, we find that the MT baseline is comparable with GPT-4, with slightly better accuracy~(14.79 vs 15.82) and slightly worse fluency~(14.60 vs 14.11). 
More importantly, when comparing these two machine translators to human translators, we observe comparable performance. 
Seamless has the lowest level of accuracy errors among all translators, and GPT-4 reaches the same level of fluency as senior translators, indicating both machine translators are reliable substitutions of human translators in resource-high directions like English$\leftrightarrow$Russian. 

In Figure \ref{fig:radar_lang}(a), we can find that each different human translator has their strengths and weaknesses, performing well or poorly in certain error categories.
For instance, Seamless has more errors in Grammar, Punctuation, and Spelling; but fewer errors in other aspects. GPT-4 is more balanced and similar to human translators in terms of errors made. 



\paragraph{English$\leftrightarrow$Chinese} 
Here, we observe a distinct pattern shift.
Seamless degrades significantly, scoring 129.19 in Accuracy and 55.46 in Fluency. 
This may be due to unbalanced training data for supervision data. 
In the meanwhile, GPT-4 is still comparable to human translators, surpassing junior/medium translators for Accuracy~(29.57 vs 35.99/37.59) for Accuracy, and performing closely for Fluency~(34.98 vs 34.19/34.18).
Senior translators substantially outperform GPT-4, attaining a score of 12.40/7.60 for Accuracy and Fluency.

From the radar chart~(Figure \ref{fig:radar_lang}(b)), we also notice a similar trend. 
Different from the English-Russian translator, Seamless has more Mistranslation, Incorrect NE, Punctuation, and Unnatural Flow errors, accounting for its performance drop. 
On the other hand, GPT-4 performs well overall, except for more errors in Incorrect NE, Grammar, and Register compared with human translators. 
As in the absence of reference, GPT-4 would translate unfamiliar words directly and literally instead of seeking online materials or other forms of help like human translators. 
For example, when translating the phrase `the Safer Transport Command', GPT4 translates the phrases to ``\begin{CJK}{UTF8}{gbsn}更(more)安全(safe)交通(traffic)指挥部(command center)\end{CJK}'' in Chinese. It implies `a more safe traffic command center' rather than capturing the intended meaning of a specialized police unit focused on transport safety.
For other errors, GPT-4 is among the level of junior/medium translators.

\begin{figure*}[t]
    \centering
    \begin{minipage}{0.32\textwidth}
        \centering
        \includegraphics[width=\textwidth]{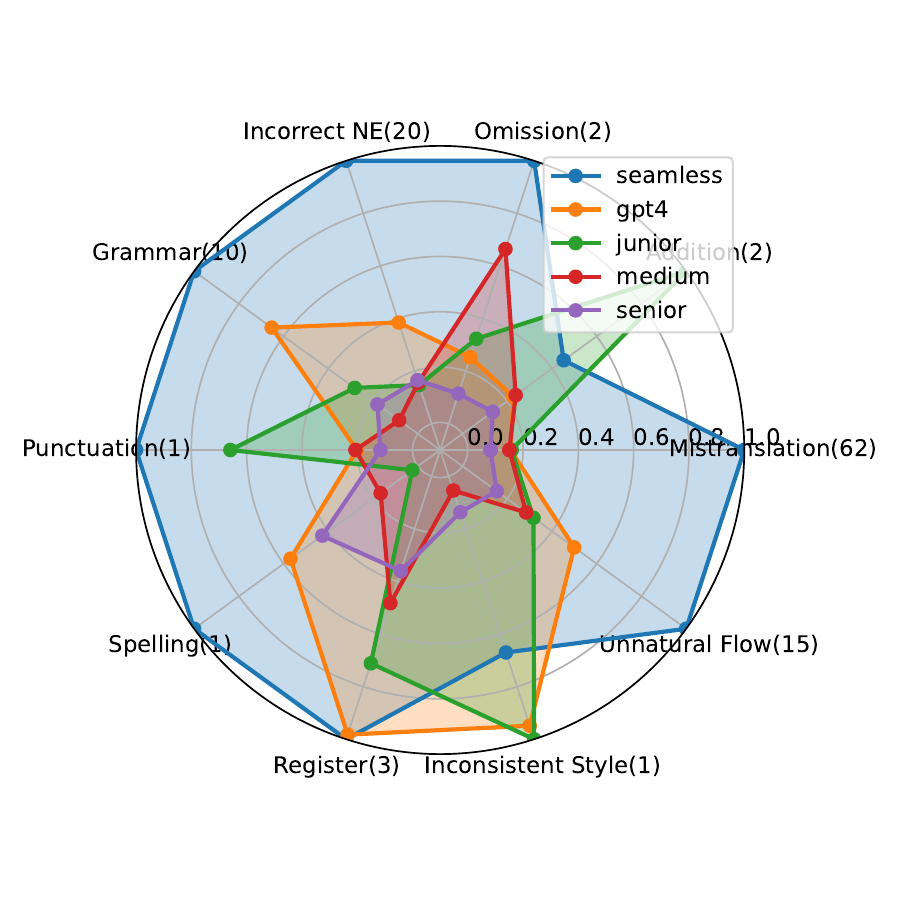}
        \caption*{(a) News domain.}
    \end{minipage}
    \hfill
    \begin{minipage}{0.32\textwidth}
        \centering
        \includegraphics[width=\textwidth]{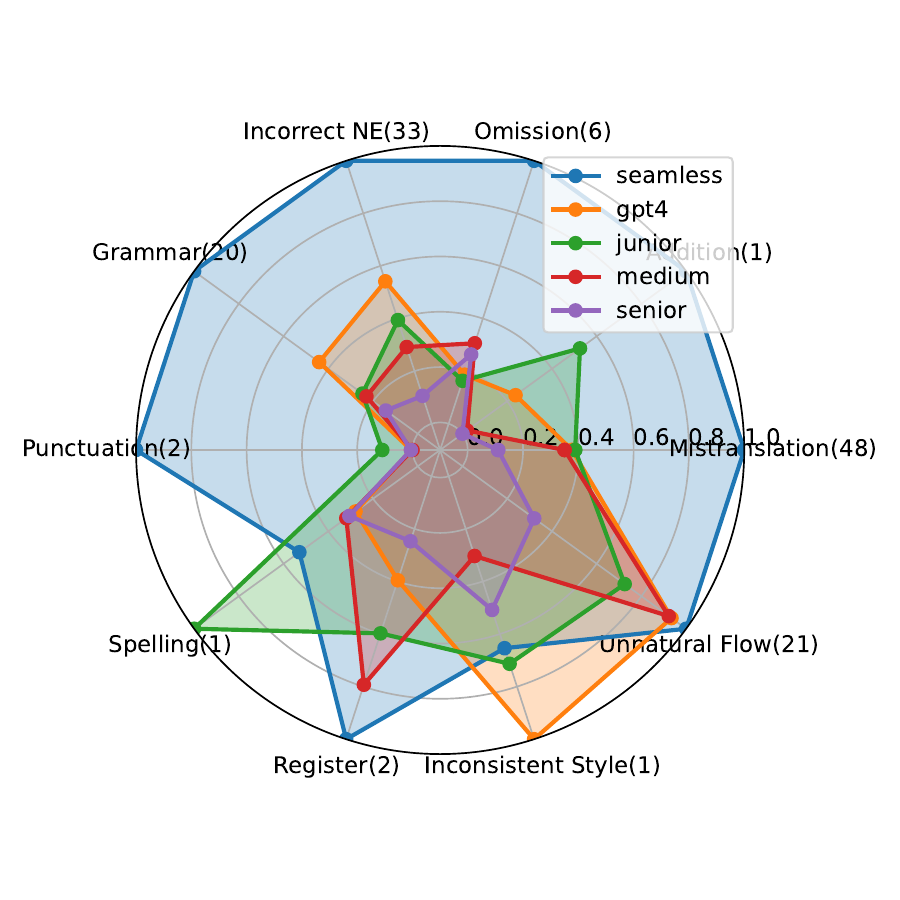}
        \caption*{(b) Technology domain.}
    \end{minipage}
    \hfill
    \begin{minipage}{0.32\textwidth}
        \centering
        \includegraphics[width=\textwidth]{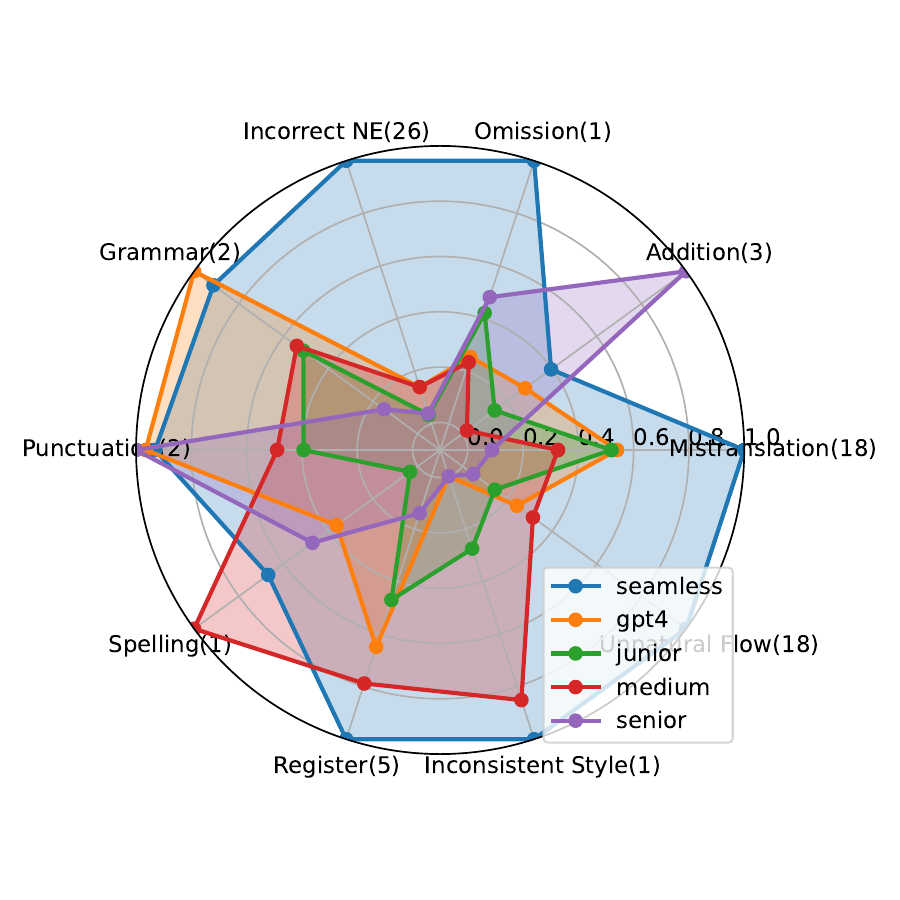}
        \caption*{(c) Biomedicine domain.}
    \end{minipage}
    \caption{Detailed errors of categories for each domain. Higher values indicate more errors and the number after each error type is the maximum number of that error.}
    \label{fig:radar_domain}
\end{figure*}

\paragraph{Hindi$\leftrightarrow$Chinese}
Compared to the directions discussed before, the overall number of errors made by all translators is higher, revealing the difficulty of our resource-low directions. 
We also notice the same pattern shift as in Chinese-English -- traditional machine translation is not reliable anymore but GPT-4 maintains its performance.
Specifically, a huge gap is observed between Seamless and other translators, with an error score over 200 vs 40-60 for accuracy, indicating an influence of scarcity of resources.
In contrast, GPT-4 attains a better Accuracy score (44.36 vs 59.03/45.39) compared with human translators.
Its fluency score~(22.83) is the same as medium human translators, and slightly weaker than senior translators~(15.19).

From Figure \ref{fig:radar_lang}(c), Seamless stands out with significantly more errors in Mistranslation~(175) than all the other translators. Unlike Seamless, GPT-4 excels in technical aspects like Spelling and Punctuation as well as maintaining source contents, demonstrated in Omission and Addition. However, GPT-4 struggles with Incorrect NE and Grammar, as in Chinese-English.



\paragraph{Discussion}

Our results here manifest an imbalance of multilinguality~\cite{wang2023not}. Our results suggest that GPT-4 is a reliable translator for all three six evaluated directions, while traditional machine translators such as Seamless are limited by the resources of the directions and only reliable in resource-high directions. 
For most cases, GPT-4 is comparable/better than junior/medium-level human translators, especially in Accuracy. 

One noticeable drawback of GPT4 across resource-medium and resource-low directions is the Incorrect NE. 
In future directions, we think web-agent could be a promising solution for resolving this issue. 
In addition, we would expect GPT4 to make a substantial number of hallucination errors as reported in various research of large language models~\cite{zhang2023siren}. While, surprisingly, we notice that in all evaluated directions, GPT4 makes almost zero hallucination errors.

\subsection{Domain-Specific Results}
\begin{table*}[t]
\centering
\begin{tabular}{|c|c|c|c|c|c|c|c|}\hline
\multicolumn{2}{|c|}{\multirow{2}{*}{\textbf{Translator}}} & \multicolumn{2}{c|}{\textbf{News}} & \multicolumn{2}{c|}{\textbf{Technology}} & \multicolumn{2}{c|}{\textbf{Biomedicine}} \\\cline{3-8}
\multicolumn{2}{|c|}{}                & Accuracy & Fluency & Accuracy & Fluency & Accuracy & Fluency \\\hline
\multirow{2}{*}{\textbf{Machine}} & Seamless & 85.38    & 26.75   & 86.58    & 43.41   & 51.70    & 26.36   \\\cline{2-8}
                         & GPT4     & 17.62    & 15.96   & 36.62    & 28.97   & 13.74    & 11.00   \\\hline
\multirow{3}{*}{\textbf{Human}}   & Junior   & 14.36    & \ph9.23    & 32.42    & 21.79   & 10.39    & \ph6.47    \\\cline{2-8}
                         & Medium   & 13.70    & \ph6.00    & 27.76    & 24.73   & \ph9.35     & 12.07   \\\cline{2-8}
                         & Senior   & \ph8.61     & \ph4.90    & 10.16    & 10.25   & \ph5.52     & \ph3.71   \\\hline
\end{tabular}
\caption{Domain Specific Results.}\label{tab:domain}
\end{table*}

Table \ref{tab:domain} and Figure \ref{fig:radar_domain} present our results for different domains in Chinese-to-English translation. 
In addition to the news domain we evaluated in the previous section, we evaluate two more domains: technology and biomedicine.  

First, in Table \ref{tab:domain}, there is a huge gap from Seamless to GPT4 and human translators, consistent with results from the previous section's English$\leftrightarrow$Chinese results. 
Despite varying levels of translation difficulty, as indicated by the average error rates among human translators, GPT-4's translation performance is marginally inferior to that of junior and mid-level translators across all three domains.
The gaps are slight for both metrics, with the notable exception of biomedical translations where GPT-4 achieves superior fluency scores.
For instance, GPT-4 scores 36.62/28.97 for the Technology domain, and human translators score [32.42, 10.16]/[24.73, 10.25]. 
Senior translators outperform GPT-4 by a noticeable margin in error numbers. 
These results are consistent with our previous findings. 
Interestingly, we find errors made by human translators are diversified~(Figure \ref{fig:radar_domain}). For instance, we observe more Unnatural Flow errors in Technology domain. Below we give more detailed discussion on each domain according to Figure \ref{fig:radar_domain}.

\paragraph{Technology Domain} 
For most of the error categories, GPT-4 reaches the level of junior and medium human translators. 
The two exceptions are Incorrect NE and Grammar, similar to our results above. 
We find that in Technology domain, there are lots of sentences with descriptive parallel phrases and in-domain named entities, which make GPT-4 struggle. Examples is presented in Section \ref{sec:case}.
Furthermore, compared to that in the News domain, there are more Unnatural Flow errors in Technology domain for both human and GPT-4 translators. 
This is due to that the information-intense and compact language style of source texts in Technology domain complicate the translation process. 

\paragraph{Biomedicine Domain} 
Unlike in Technology, we notice no clear disadvantage of Incorrect NE for GPT-4 compared to human translators. 
In addition, there are fewer Grammar errors for Biomedicine domain. 
Human translators occasionally demonstrate specific weaknesses. For example, here we find the senior translator makes more Addition errors. Meanwhile, our medium translator has more Register and Inconsistent Style errors. 
These errors are not systematic; rather, each translator exhibits its own distinct weaknesses. In GPT-4's case, the primary limitations manifest in Named Entity recognition and grammatical accuracy.

In summary, for our evaluated domains, we show that GPT-4 is a reliable translator as junior/medium human translators, except for mild Incorrect NE and Grammar issues. 
The relative performance comparing with human translators remain consistent across domains. 



\begin{table*}[]
\centering
\begin{tabular}{|c|c|p{10cm}|}\hline
\multirow{3}{*}{Named-Entity} & Source & \begin{CJK}{UTF8}{gbsn}巨人\underline{网络}有限公司\end{CJK} \\\cline{2-3}
& GPT-4 &  Giant {\color{red}Network} Group Inc.\\\cline{2-3}
& Human & Giant {\color{blue}Interactive} Group Inc.\\\hline\hline
\multirow{3}{*}{Unnatural Flow} & Source & \underline{It's just a white screen} or it times out loading it, or the page becomes unresponsive! \\\cline{2-3}
& GPT-4 & \begin{CJK}{UTF8}{gbsn}{\color{red}它只是一个白屏}，要么是加载时超时，要么页面变得无响应了！\end{CJK} \\\cline{2-3}
& Human & \begin{CJK}{UTF8}{gbsn}{\color{blue}页面要么显示空白}，要么加载超时或是无响应。\end{CJK} \\\hline\hline
\multirow{3}{*}{Consistent Usage} &Source &  \begin{CJK}{UTF8}{gbsn}
最后再来看下仪表盘和座椅，仪表盘设计可圈可点，设计比较稳重。该车采用了仿皮座椅，\underline{座椅包裹性到位，用料讲究，乘坐舒适}。
\end{CJK}\\\cline{2-3}
&GPT-4 & Finally, let's take a look at the dashboard and seats. The dashboard design is noteworthy and quite stable. The car features faux leather seats that {\color{red}are well-contoured, made with quality materials, and offer a comfortable ride.} \\\cline{2-3}
&Human & Finally, let's take a look at the dashboard and the seat, the dashboard is designed to be circular, the design is more stable, the car uses a leather seat, {\color{blue}the seat is wrapped in place, the material is careful, and the ride is comfortable.}\\\hline\hline
\multirow{3}{*}{Human Hallucination} &Source &  He has health concerns atm but \underline{we also have Daley entering his 2nd year} and is a decent safety net.\\\cline{2-3}
&GPT-4 &  \begin{CJK}{UTF8}{gbsn}他目前有健康问题，但我们还有戴利进入他的第二年，{\color{red}他是一个不错的安全保障。(literal)}\end{CJK}\\\cline{2-3}
&Human & \begin{CJK}{UTF8}{gbsn}他目前有健康问题。不过，{\color{red}戴利两岁了}，是个不错的备选人。\end{CJK} \\\hline
\end{tabular}
\caption{Case study. Underline indicates the critical phrases in the source sentence, text in red indicates errors, and text in blue indicates correct usage.}\label{tab:cases}
\end{table*}

\subsection{Case Study}\label{sec:case}
We also qualitatively understand the difference between the translations given by GPT-4 and human translators. 
\paragraph{Literal Translations}
Among the error cases, the typical one is literal translations. Specifically, we find that GPT-4 sometimes translates with semantically correct, but in-native and literal translations. 
This is problematic with named entities, especially those occurring less frequently. 
As shown in Table \ref{tab:cases}, when not knowing the correct translation of \begin{CJK}{UTF8}{gbsn}`巨人(Giant)网络(Network)有限公司(Limited Company)'\end{CJK}, GPT-4 translates the term word by word. 
However, the issue of named entities occurs less for human translators, partially because they would google it to find the correct translation. Thus, this issue might be resolved by incorporating web-search into agent-like translation~\cite{mt-agent-1,mt-agent-2}. 

Except for named entities, we notice that the literal translation causes Unnatural Flows. 
As shown in Table \ref{tab:cases}, when translating `It's just a white screen', GPT-4 translates the phrase to \begin{CJK}{UTF8}{gbsn}`它(it)只是(is just)一个(a)白屏(white screen)'\end{CJK}, but human translator translates this phrase to `\begin{CJK}{UTF8}{gbsn}`页面显示空白(The page display is white)'\end{CJK}', which represents a more precise meaning and follows local conventions.

\paragraph{Consistent Word Usage}
In the previous section, we observed that GPT4 sometimes makes more Grammar errors. By delving into cases, we find that GPT4 occasionally makes grammar errors when involving consistent usage of words or phrases. For example in Table \ref{tab:cases}, GPT4 makes a faulty parallelism error, i.e., elements in a series do not match. In this example, `well-contoured'~(adjective), `made with quality materials'~(passive phrase), and `offer a comfortable ride'~(active verb phrase) are not in parallel form. 
In other cases, GPT-4 sometimes uses inconsistent pronouns or tenses, e.g., `is' before and `was' after.

\paragraph{Human Hallucination}
We find human translators also have drawbacks compared to the GPT-4 translator. 
When the source sentence contains insufficient information to translate, human translators tend to fill the gap by imagination or overthinking. 
An example is given in Table \ref{tab:cases}. 
The translator incorrectly understands the phrase `entering his 2nd year' as Daley is a two-year-old baby, but the sentence describes a 2nd-year player for sports. 
This may be due to daily language habits, misunderstanding, or not spending sufficient time on reasoning.
The mistakes can be inevitable given the nature of a professional translator's work and could be related to the hallucination~\cite{zhang2023siren} of LLMs. 
GPT-4's literal translation helps in this, as it keeps faithful to the source sentence. This also aligns with our findings in Section \ref{sec:overall} that GPT-4 has fewer Additions or Omissions.







\section{Conclusion}
We comprehensively evaluated the translation quality of GPT-4 against human translators of varying expertise levels across multiple language pairs and domains, finding that the LLM performs significantly better than the Seamless, a traditional NMT system, and comparably to junior/medium translators in terms of total errors made. However, LLMs still lag behind senior translators. We also notice that GPT-4's translation capability weakens from resource-rich to resource-poor language pairs. Qualitative analysis revealed that GPT-4 produces more literal translations and consistent word usage than human translators but suffers less from imagined information.
The results above demonstrate that in all directions, LLMs have the potential to replace human translators, especially junior and medium ones feasibly. To our knowledge, we are the first to report evaluation of machine translation systems by ranking them among professional human translators according to industry-level metrics.

\bibliographystyle{IEEEtran}
\bibliography{custom}

\appendix
\section{Expertise of Human Annotators}
\label{sec:expertise}
To categorize translators into junior, medium, or senior levels, we have established a comprehensive set of criteria that take into account various factors indicative of a translator's expertise and experience. These factors include the translator's educational background, particularly the prestige of the institution from which they graduated, as well as their length of service in the translation industry, the duration of their translation career, the number of translations completed, and any professional certifications they have obtained. To ensure the ongoing competence of our translators, we conduct quarterly assessments to evaluate their performance. For instance, to be classified as a senior-level translator, an individual must possess a minimum of ten years of translation experience, demonstrate exceptional proficiency by achieving a score of 99\% on our assessments, and hold the distinguished CATTI++ translation certification. By considering these stringent criteria, we aim to maintain a highly qualified and skilled pool of translators across all levels of expertise.

\section{Annotation Requirements}

\subsection{Error Types}

\begin{figure*}
    \centering
    \includegraphics[width=1.0\textwidth]{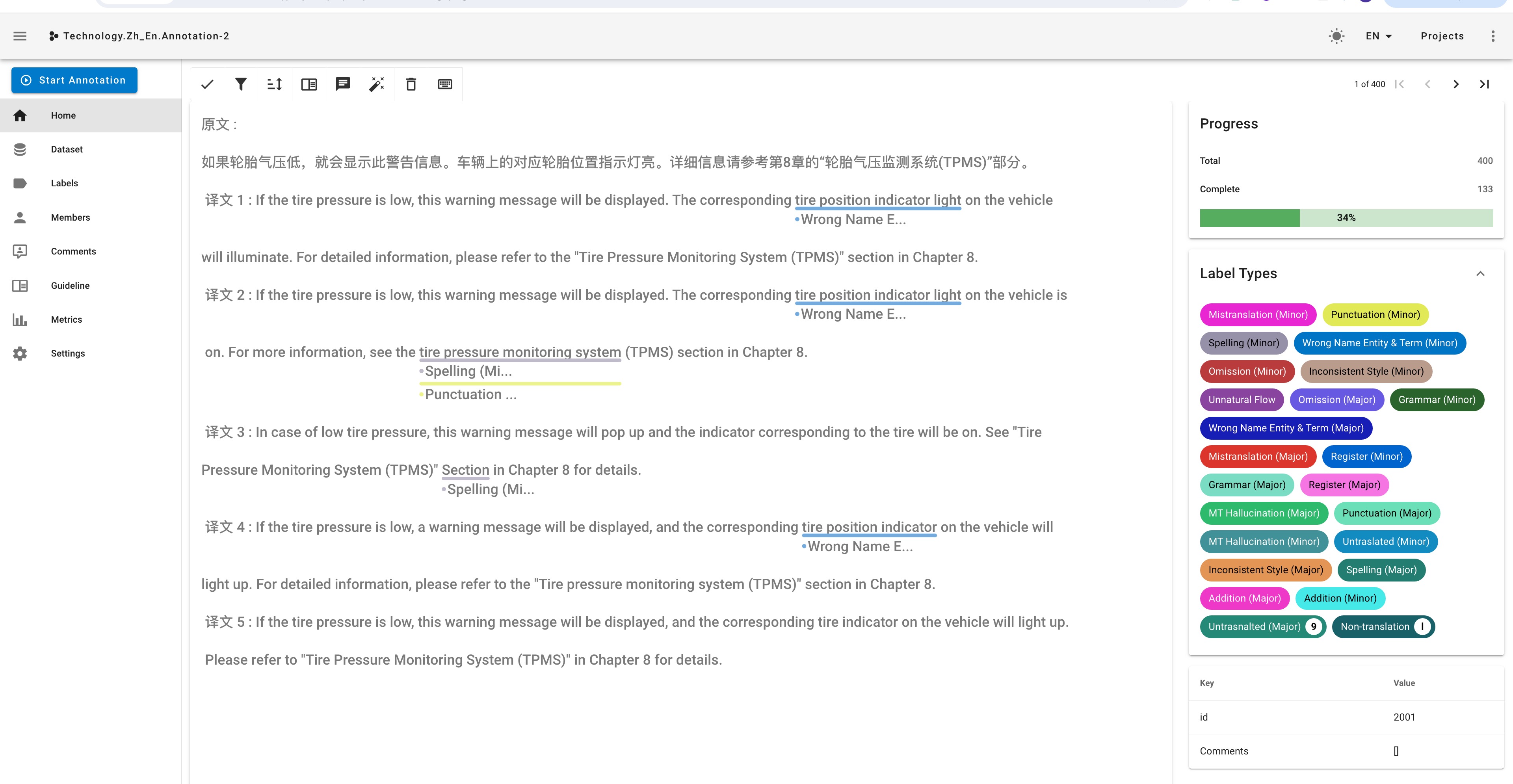}
    \caption{A screenshot of the Doccano annotation system we use. }
    \label{fig:screenshot}
\end{figure*}

Our annotation system is built upon the open-sourced doccano system~\footnote{\url{https://github.com/doccano/doccano}}. 
In Figure \ref{fig:screenshot}, we provide a screenshot of our annotation system. 
For each source sentence, outputs for different systems are given and the annotators can select spans of the text and annotate the error type and severity. 

\section{Detailed Explanation and Guidance for Each Error Types}
Our evaluation protocol largely follows the MQM criteria released by Unbabel\footnote{}. 
We provide a detailed annotation manual for annotators, including an explanation for each error type as well as illustrative examples for error types. 
It is included in the following:


\subsection{Annotation Requirements}
    The minimum unit that can be selected and annotated is a whole word, a whitespace, a punctuation mark, or an isolated character. 
    In the following example, the version in French has an extra exclamation mark, so it's necessary to annotate it as a Punctuation error:
    
    
    \emph{[EN] Thank you very much.} 
    
    \emph{[FR] Merci beaucoup!} 
    
    \begin{mdframed}
        Wrong selection $\rightarrow$Merci [beaucoup!]PUNCTUATION
        
        Correct selection $\rightarrow$Merci beaucoup[!]PUNCTUATION
    \end{mdframed} 
    
    If the issue occurs in a multiword expression, you will need to select the whole expression; if, for example, an entire sentence was translated and it shouldn't have been, you should select the entire sentence. 
    
    In the following example, we have an Unnatural Flow error: 
    
    \noindent\emph{[EN] Hi, Mary here.} 
    
    \noindent\emph{[ES] Hola, Mary aqu\'{i}.} 
    
    \begin{mdframed}
        Wrong selection $\rightarrow$ Hola, [Mary aqu\'{i}.]UNNATURAL FLOW 
        
        Correct selection $\rightarrow$ Hola, [Mary aqu\'{i}]UNNATURAL FLOW.
    \end{mdframed}

\subsection{Error Types}
\paragraph{Accuracy} 
            \begin{itemize}
                \item Mistranslation
                \begin{itemize}
                    \item Description: Translation does not accurately represent the source.
                    \item Example: 
                    \begin{mdframed}
                        \emph{[EN] It has to be done by the book.} 
                        
                        \emph{[FR] Il doit \^{e}tre fait [par le livre]MISTRANSLATION} 
                        
                        \emph{[Reason] The word-for-word translation into French doesn't work.} 
                    \end{mdframed}
                \end{itemize}
                \item Addition
                \begin{itemize}
                    \item Description: Information not present in the source.
                    \item Example: 
                    \begin{mdframed}
                        \emph{[EN] That way you can be sure that you were the one who made the changes.} 
                        
                        \emph{[ES] As\'{i} puedes estar seguro de que fuiste t\'{u} quien hizo [todos ADDItIoN los cambios.} 
                        
                        \emph{[Reason] [Todos] (meaning 'all' in Spanish) is not present in the source and it is incorrectly added in the target text.}
                    \end{mdframed}
                \end{itemize}
                \item MT Hallucination
                \begin{itemize}
                    \item Description: information that has nothing related to source; or gibberish; or repeats 
                    
                    \item Example:
                    \begin{mdframed}
                        \emph{[EN] You can send us a follow-up email at this address [EMAIL].} 
                        
                        \emph{[ES] [H\'{a}game saber si tiene alguna otra pregunta]MT HALLUCINATION.]} 
                        
                        \emph{[Reason]: The Spanish translation reads please let me know if you have any other questions and it's grammatically correct and fluent, but it has no relation at all with the source.]}
                    \end{mdframed}
                \end{itemize}
                \item Omission
                \begin{itemize}
                    \item Description: Missing content from the source.
                    \item Example:
                    \begin{mdframed}
                        \emph{[EN] We do not have much information on this.} 
                        
                        \emph{[FR] Nous ne disposons pas [] OMISSION beaucoup d'informations \`{a} ce sujet.} 
                        
                        \emph{[Reason]: The French sentence requires the preposition [de] (disposer de).}
                    \end{mdframed}
                \end{itemize}
                \item Untranslated
                \begin{itemize}
                    \item Description: Not translated.
                    \item Example:
                    \begin{mdframed}
                        \emph{[EN] How To Make Pizza Dough} 
                        
                        \emph{[FR] Comment faire de [Pizza Dough|UNTRANSLATED} 
                        
                        \emph{[Reason]: [Pizza Dough] is not a named entity and is untranslated in the French version.}
                    \end{mdframed}
                \end{itemize}
                \item Wrong Name Entity \& Term
                \begin{itemize}
                    \item Description: Wrong usage of NE and Terminology.
                    \item Example:
                    \begin{mdframed}
                        \emph{[EN] Dear Wiley, } 
                        
                        \emph{[IT] Gentile [Wilar WRONG NAMED ENTITY, } 
                        
                        \emph{[Reason]: The name in the Italian version doesn't match the original.}
                    \end{mdframed}
                \end{itemize}
            \end{itemize}
        \paragraph{Fluency} 
        \begin{itemize}
            \item Grammar
            \begin{itemize}
                \item Description: Problems with grammar of target language.
                \item Example:
                \begin{mdframed}
                    \emph{[EN] I understand that you want to check in online.} 
                    
                    \emph{[CS] ch\`{a}pu, ze se chcete [odbaven\'{i}]gRAMMaR online.} 
                    
                    \emph{[Reason]: Wrong part of speech makes the sentence ungrammatical in Czech.}
                \end{mdframed}
            \end{itemize}
            \item Punctuation
            \begin{itemize}
                \item Description: incorrect punctuation (for locale or style). 
                \item Example:
                \begin{mdframed}
                    \emph{[EN] Original copy of the Proof of Purchase or Invoice (not a screenshot): } 
                    
                    \emph{[PT] C'{o}pia original do comprovante de compra ou nota fiscal (n\~{a}o uma captura de tela)[.]PUNCTUATION} 
                    
                    \emph{[Reason]: There's a period instead of a colon in the Brazilian Portuguese version of this sentence.}
                \end{mdframed}
            \end{itemize}
            \item Spelling
            \begin{itemize}
                \item Description: incorrect spelling or capitalization.
                \item Example:
                \begin{mdframed}
                    \emph{[EN] This sort of damage is not covered under the warranty, but we will seek assistance from a higher support and see what we can do regarding this issue.} 
                    
                    \emph{[IT] Questo tipo di danno non \`{e} coperto dalla garanzia, ma chieder\`{o} comunque aiuto ai responsabili dell'assistenza per capire che cosa [Zi]SPELLING pu\`{o} fare per quanto riguarda questo problema.} 
                    
                    \emph{[Reason]: There's a typo in the sentence in Italian: the word [zi] should be [si] instead.}
                \end{mdframed}
            \end{itemize}
            \item Register
            \begin{itemize}
                \item Description: Wrong grammatical register (e.g., inappropriately informal pronouns).
                \item Example:
                \begin{mdframed}
                    \emph{[EN] Wishing you a great day ahead.} 
                    
                    \emph{[DE] Ich w\"{u}nsche [Ihnen]REGISTER einen sch\"{o}nen Tag.} 
                    
                    \emph{[Reason]: The required register for the German translation is Informal but the pronoun [Inhen] is Formal.}
                \end{mdframed}
            \end{itemize}
            \item Inconsistent Style
            \begin{itemize}
                \item Description: internal inconsistency (not related to terminology). 
                
                \item Example:
                \begin{mdframed}
                    \emph{[EN] Please click on this link. [...] This link will expire in 24 hours.} 
                    
                    \emph{[NN] Klikk p\r{a} denne [lenken].[...]Denne [linken]INCONSISTENCY utloper om 24 timer.} 
                    
                    \emph{[Reason]: Both [lenk] and [link] are correct in Norwegian, but in the same document, only one should be used. Note: this is a single error, not two}
                \end{mdframed}
            \end{itemize}
            \item Unnatural Flow
            \begin{itemize}
                \item Description: translations that are too literal or sound unnatural.
                \item Example:
                \begin{mdframed}
                    \emph{[EN] Zebras are ideal for animal matching.} 
                    
                    \emph{[DE] [Zebras sind ideal, um bestimmte Tiere zu finden]UNNATURAL FLOW.} 
                    
                    \emph{[Reason] The German translation sounds too literal, it reads like a translation, using the verb [finden] (finding) as a translation for matching. The verb matching should be translated as [detektieren] (detect) to read as if it was originally written in the target language: [Zebras sind ein ideales Beispiel zur Detektion von Wildtieren.]}
                \end{mdframed}
            \end{itemize}
        \end{itemize}
\paragraph{Other} 
        \begin{itemize}
            \item Non-translation
        \end{itemize}





\vfill

\end{document}